\documentclass[sigconf]{acmart}
\usepackage{multirow}
\usepackage{graphicx}  
\usepackage{subcaption}

\AtBeginDocument{%
  }

\renewcommand\footnotetextcopyrightpermission[1]{}
\begin{document}

\title{EBAD-Gaussian: Event-driven Bundle Adjusted Deblur Gaussian Splatting}

\author{Yufei Deng}
\affiliation{%
  \institution{Sichuan University}
  \city{Chengdu}
  \country{China}
}
\email{yufeideng@stu.scu.edu.cn}

\author{Yuanjian Wang}
\affiliation{%
  \institution{Sichuan University}
  \city{Chengdu}
  \country{China}
}
\email{wangyuanjian@stu.scu.edu.cn}

\author{Rong Xiao}
\affiliation{%
  \institution{Sichuan University}
  \city{Chengdu}
  \country{China}
}
\email{rxiao@scu.edu.cn}

\author{Chenwei Tang}
\affiliation{%
  \institution{Sichuan University}
  \city{Chengdu}
  \country{China}
}
\email{tangchenwei@scu.edu.cn}

\author{Jizhe Zhou}
\affiliation{%
  \institution{Sichuan University}
  \city{Chengdu}
  \country{China}
}
\email{jzzhou@scu.edu.cn}

\author{Jiahao Fan}
\affiliation{%
  \institution{Sichuan University}
  \city{Chengdu}
  \country{China}
}
\email{fanjh@scu.edu.cn}

\author{Deng Xiong}
\affiliation{%
  \institution{Stevens institute of Technology}
  \city{Hoboken}
  \state{NJ}
  \country{China}
}
\email{dxiong@stevens.edu}

\author{Jiancheng Lv}
\affiliation{%
  \institution{Sichuan University}
  \city{Chengdu}
  \country{China}
}
\email{lvjiancheng@scu.edu.cn}

\author{Huajin Tang}
\affiliation{%
  \institution{Zhejiang University}
  \city{Hangzhou}
  \country{China}
}
\email{htang@zju.edu.cn}

\begin{abstract}
While 3D Gaussian Splatting (3D-GS) achieves photorealistic novel view synthesis, its performance degrades with motion blur. In scenarios with rapid motion or low-light conditions, existing RGB-based deblurring methods struggle to model camera pose and radiance changes during exposure, reducing reconstruction accuracy. Event cameras, capturing continuous brightness changes during exposure, can effectively assist in modeling motion blur and improving reconstruction quality. Therefore, we propose Event-driven Bundle Adjusted Deblur Gaussian Splatting (EBAD-Gaussian), which reconstructs sharp 3D Gaussians from event streams and severely blurred images. This method jointly learns the parameters of these Gaussians while recovering camera motion trajectories during exposure time. Specifically, we first construct a blur loss function by synthesizing multiple latent sharp images during the exposure time, minimizing the difference between real and synthesized blurred images. Then we use event stream to supervise the light intensity changes between latent sharp images at any time within the exposure period, supplementing the light intensity dynamic changes lost in RGB images. Furthermore, we optimize the latent sharp images at intermediate exposure times based on the event-based double integral (EDI) prior, applying consistency constraints to enhance the details and texture information of the reconstructed images. Extensive experiments on synthetic and real-world datasets show that EBAD-Gaussian can achieve high-quality 3D scene reconstruction under the condition of blurred images and event stream inputs.

\end{abstract}

\ccsdesc[500]{Computing methodologies~Reconstruction}

\keywords{3D Reconstruction, Event Camera, Motion Blur Removal, Multi-modal Fusion}

\maketitle

\section{Introduction}
Recent advances in 3D reconstruction and novel view synthesis have driven the development of key applications, including augmented and virtual reality (AR/VR) \cite{tang2024cycle3d, yuen2011augmented}, robot navigation \cite{gul2019comprehensive, deng2023semantic}, scene understanding \cite{guo2024semanticgaussiansopenvocabularyscene,zhou2024hugsholisticurban3d}, and 3D content generation \cite{dai2024gonerfgeneratingobjectsneural, lin2023magic3d}. Among these, 3D Gaussian Splatting (3D-GS) \cite{kerbl20233dgaussiansplattingrealtime, chen_3dgs_survey_2024} can generate high-fidelity novel views with high rendering speed, provided there are high-quality 2D images and accurate camera poses. However, in practical applications, obtaining high-quality images and precise poses can be challenging. Image clarity is influenced by both the exposure time and the camera's motion speed. During exposure, the sensor continuously records light. If the camera or object moves rapidly, the same object will be imaged at different locations on the sensor, causing its trajectory to be captured as motion blur. The longer the exposure time and the faster the camera moves, the more pronounced the object's motion blur becomes, resulting in a blurrier image. Blurry images lead to feature point matching errors, increasing the deviation of camera poses and point clouds estimated by Structure from Motion (SfM) \cite{schonberger2016structure}, while the loss of details affects Gaussian parameter optimization, ultimately reducing the quality of novel view synthesis. Therefore, under conditions of fast camera motion and low light, how to recover sharp 3D-GS scenes from motion-blurred images becomes a key issue.

In practical applications, when the scene remains static, motion blur arises from camera pose changes during exposure. The key challenge in reconstructing sharp 3D-GS is accurately recovering the motion trajectory to simulate blur formation. Recently, BAD-Gaussians~\cite{zhao2024badgaussiansbundleadjusteddeblur} modeled the physical image formation process of motion-blurred images and jointly optimized the camera motion trajectory and Gaussian parameters during exposure, enabling the recovery of high-quality 3D scenes from blurred images. However, this method is only applicable to scenes where blur is caused by global camera motion and struggles under conditions of low lighting and fast motion. This is due to the low signal-to-noise ratio of images, which leads to increased errors in camera trajectory estimation under low-light conditions; meanwhile, fast motion introduces stronger spatially inconsistent blur, making the global trajectory modeling assumption invalid. Therefore, under these extreme conditions, it remains challenging for this method to recover fine details in the reconstructed scene.

Event cameras provide a more effective solution for the deblurring task in low-light and fast-motion scenarios~\cite{qi2023e2nerf, cannici2024mitigatingmotionblurneural, Qi_2024, klenk2023enerfneuralradiancefields}. As a bio-inspired visual sensor, they asynchronously detect brightness changes in the scene, capturing subtle dynamic information with high temporal resolution, thus compensating for the key information lost during the exposure time in RGB cameras \cite{Gallego_2022}. Inspired by this, E2NeRF \cite{qi2023e2nerf} introduces event information, using consistency loss and structure-aware constraints to guide the neural radiance field (NeRF) \cite{mildenhall2020nerfrepresentingscenesneural} in capturing spatiotemporal structural changes, achieving scene reconstruction without sharp image supervision. Ev-DeblurNeRF \cite{cannici2024mitigatingmotionblurneural} further utilizes event streams to model the motion blur process, optimizing the NeRF by combining event information and RGB data, thereby improving the reconstruction quality. Compared to methods relying solely on RGB cameras, these methods can recover more accurate details. However, achieving real-time rendering and synthesizing high-fidelity novel views with complex details presents a significant challenge for these approaches.

To address the above challenges, this paper proposes the EBAD-Gaussian method, which effectively suppresses motion blur and achieves high-quality real-time reconstruction by fusing event streams and RGB bimodal data, and optimizing Gaussian parameters and camera poses within the exposure time during training. Specifically, we first model the cumulative effect of relative camera motion over the exposure time and static scene on imaging. By synthesizing multiple latent sharp images, we construct a blur loss function to minimize the difference between real blurred images and synthesized blurred images, providing a foundational guarantee for the clarity and structural fidelity of the reconstructed images. Next, we simulate the predicted event data by modeling the brightness change during the exposure time and compare it with the real event stream, serving as the event rendering loss. Additionally, we compare the event-based double integral (EDI) \cite{pan2018bringingblurryframealive} deblurred image with the latent sharp image at intermediate exposure moments to construct the EDI loss. By jointly constraining the 3D-GS training process with the above loss functions, EBAD-Gaussian not only effectively removes motion blur but also significantly improves reconstruction accuracy and real-time rendering capability. Experiments on both synthetic and real data demonstrate that the proposed method achieves high-fidelity real-time reconstruction under complex motion and low-light conditions, providing an efficient solution for the 3D reconstruction of static scenes. In summary, we make the following contributions:
\begin{itemize}
    \item To address the limitations of traditional 3D-GS in low-light and high-speed scenarios, we leverage the high dynamic range characteristics of event streams to compensate for high-frequency details in underexposed regions. 
    \item We propose a novel EBAD-Gaussian framework that incorporates a motion blur loss and jointly optimizes camera poses and Gaussian parameters using both event and RGB data, thereby significantly enhancing the reconstruction quality.
    \item Experiments on both synthetic and real data validate that our method achieves precise motion trajectory estimation and high-fidelity real-time 3D reconstruction for scenes with severe blur images and corresponding event data.
\end{itemize}

\section{Related Work}
\subsection{Recovering 3D Structure from Blurred Images}
In high-quality 3D scene reconstruction, sharp and high-fidelity images are crucial as supervision signals. However, motion-blurred images, commonly encountered in the real world, can severely impact the accurate reconstruction of 3D scenes. Recent advances in neural rendering have addressed the motion blur problem through distinct methodological approaches. In the seminal work of Deblur-NeRF \cite{ma2022deblurnerf} and DP-NeRF \cite{Lee_2023_CVPR_DPNeRF}, the image degradation process is modeled via learned blur kernels that approximate the integration of radiance over the exposure period. Building upon this framework, BAD-NeRF \cite{wang2023badnerf} introduces a physically-grounded blur formation model, where the implicit neural representation is jointly optimized with the camera trajectory through a differentiable bundle adjustment formulation. This co-optimization paradigm simultaneously refines both the scene geometry and the temporally-varying camera poses during the exposure interval.
However, these NeRF-based methods suffer from high computational costs and long training times, making it difficult for them to meet the demands of real-time applications. Building upon recent advancements in 3D-GS \cite{ye2024gsplatopensourcelibrarygaussian, kheradmand20243d}, BAD-Gaussians \cite{zhao2024badgaussiansbundleadjusteddeblur}proposed a novel framework that integrates 3D-GS scene representation with physically-based blur modeling and bundle adjustment for deblurring reconstruction. This approach demonstrates superior computational efficiency compared to neural radiance field methods, achieving both faster rendering speeds and accelerated convergence. Nevertheless, significant challenges remain in processing heavily degraded images captured under extreme conditions of low illumination and rapid motion.
\subsection{Event-Based 3D Scene Reconstruction}
The event camera is a novel bio-inspired visual sensor. Unlike standard cameras that capture images at a fixed frame rate, event cameras generate an event stream by detecting changes in pixel brightness. They are characterized by low latency, high dynamic range, low power consumption, and high temporal resolution. Inspired by the excellent performance of event cameras, several studies have attempted to reconstruct 3D scenes from the event stream captured by event cameras, especially under conditions of low light and rapid camera motion. Recently, various event-based 3D reconstruction methods\cite{xiong2024event3dgs, chakravarthi2024recent, han2024event, rudnev2024dynamic} have been proposed, including EventNeRF \cite{rudnev2022eventnerf}, Ev-NeRF \cite{Hwang_2023_WACV}, SaENeRF \cite{wang2025saenerf}, Robust e-NeRF \cite{low2023_robust-e-nerf} and its variants \cite{low2024_deblur-e-nerf}. These methods focus on reconstructing 3D scenes from the event streams generated by fast-moving event cameras. However, reconstructions relying solely on event data often suffer from limited texture quality. 

E-NeRF\cite{klenk2023e-nerf} was the first to utilize both event streams and intensity images for 3D reconstruction, enabling sharper radiance field generation. Nevertheless,this method does not model the formation of blurred images, limiting its reconstruction fidelity. To address this, Ev-DeblurNeRF \cite{cannici2024mitigatingmotionblurneural} incorporated a blur formation model and leveraged event data to supervise brightness variations during camera exposure, leading to improved reconstruction results. However, it does not optimize camera poses during training, which restricts its performance. EBAD-NeRF\cite{qi2024deblurring} further optimizes the reconstruction quality by jointly optimizing the camera pose and implicit representation of the scene throughout the exposure time. Yet, the high computational cost of ray sampling in neural radiance field frameworks remains a significant bottleneck, limiting the real-time applicability of such methods in practical scenarios.
\begin{figure*}[htbp]
    \centering
    \includegraphics[width=0.9\linewidth]{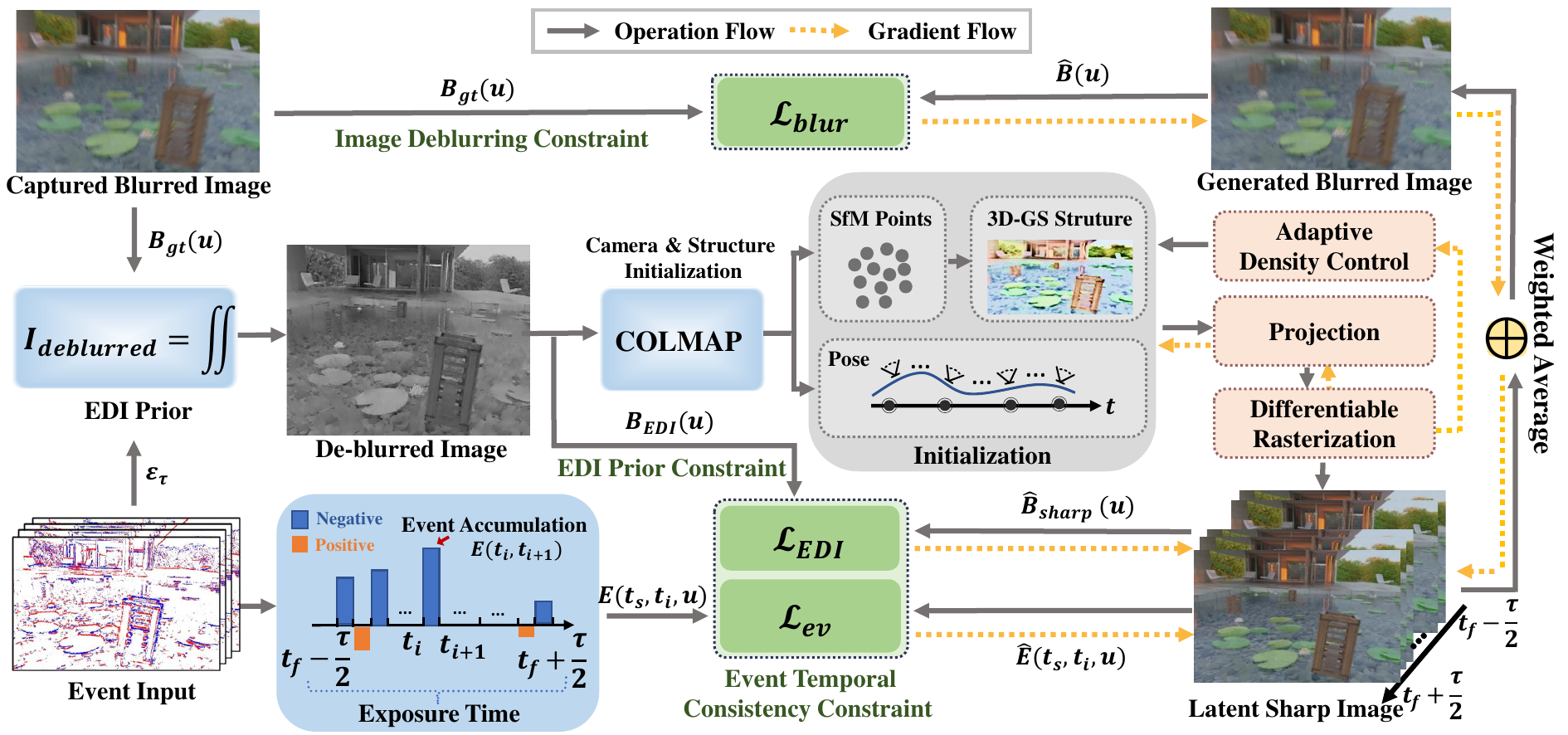}  
    \caption{Overview of the proposed EBAD-Gaussian framework. The method jointly optimizes the 3D Gaussian parameters and camera poses during exposure, using motion blur constraints, event supervision, and EDI priors to achieve high-fidelity scene reconstruction.}
    \Description{This figure shows an overview of the proposed EBAD-Gaussian framework. It illustrates how the method optimizes both the 3D Gaussian parameters and camera poses, incorporating motion blur constraints, event supervision, and EDI priors for high-fidelity scene reconstruction.}
    \label{fig:framework}
\end{figure*}
\section{Methods}
EBAD-Gaussian leverages dual-modal data, namely images with motion blur and event data, to jointly optimize Gaussian parameters and camera poses during the exposure time, enabling the reconstruction of sharp and high-quality 3D Gaussian splatting.
Specifically, we first use the EDI \cite{pan2019bringing} algorithm to process the blurred images and the corresponding event stream to generate the deblurred latent sharp image. Next, the EDI deblurred image is input into COLMAP \cite{schoenberger2016sfm} to generate an SfM sparse point cloud and an initial camera pose recovered from the motion. Then, we initialize the 3D-GS \cite{kerbl20233dgs} scene representation and project it onto the camera imaging plane, rasterizing to generate the latent sharp image. Finally, the average of the latent sharp images is utilized to simulate motion blur, and the loss is constructed by comparing it to the real blurred image; the EDI deblurred image constraint ensures consistency between the latent sharp image captured at intermediate moments during the exposure time; and the constraint imposed by the integral of the event stream, when multiplied by a threshold, regulates the differences in the logarithmic domain of brightness between the latent sharp images at different moments during the exposure time. Through these three constraints, we can effectively optimize the Gaussian parameters and camera pose, achieving high-fidelity modeling of motion blur and generating high-quality dynamic scene reconstruction results (see Figure~\ref{fig:framework}). 

\subsection{Motion Blur Modeling with 3D-GS}
3D Gaussian Splatting models static scenes using an explicit set of Gaussian primitives. Each Gaussian primitive is characterized by its center position $\mu \in \mathbb{R}^3$, covariance matrix $\Sigma \in \mathbb{R}^{3 \times 3}$, opacity $\alpha \in [0, 1]$, and third-order spherical harmonic coefficients, which are used to represent its spatial structure, color attributes, and directional dependence. During the rendering process, 3D-GS projects each Gaussian primitive from the world coordinate system onto the camera's image plane, and the projection covariance on the image plane can be computed using the following formula:
\begin{equation}
\Sigma' = J W \Sigma W^{\top} J^{\top}
\end{equation}
Where $W$ is the transformation matrix representing the change from the world coordinate system to the camera coordinate system, and $J$ is the Jacobian approximation of the perspective projection. Subsequently, the system sorts all the Gaussians based on depth and performs image composition using volume rendering\cite{max1995optical, kajiya1984ray}. The final color at the pixel position $u = (x, y)$ is given 
:
\begin{equation}
C(u) = \sum_{i=1}^{n} c_i \alpha_i \prod_{j=1}^{i-1} (1 - \alpha_j)
\end{equation}
Where $c_i$ is the color of the $i$-th Gaussian and $\alpha_i$ is its opacity. Since this process is fully differentiable, it supports optimization based on image supervision.

The imaging process of modern digital cameras is essentially the spatiotemporal integration of photons over a finite exposure time. When there is relative motion between the camera and the scene being captured, the radiance value recorded at each pixel $u$ on the sensor can be modeled as the temporal integration of the latent sharp image over the exposure time window $[f - \tau/2, f + \tau/2]$:
\begin{equation}
B(u) = \frac{1}{\tau} \int_{f-\frac{\tau}{2}}^{f+\frac{\tau}{2}} I(u,t) \, dt
\label{model_blur}
\end{equation}
Where $B(u) \in \mathbb{R}^{W \times H \times 3}$ represents the observed blurry image, with tensor dimensions corresponding to the image width $W$, height $H$, and the RGB channels of radiance. $I(u,t) \in \mathbb{R}^{W \times H \times 3}_+$ represents the radiance received at the pixel $u$ on the sensor plane at time $t$ under ideal non-blurry conditions. $\tau$ is the total exposure time, and $f$ is the middle exposure time, satisfying:
\begin{equation}
f = \frac{t_{\text{start}} + t_{\text{end}}}{2}
\end{equation}
where $t_{\text{start}}$ and $t_{\text{end}}$ are the start and end times of the camera exposure, respectively. $\frac{1}{\tau}$ is the normalization factor used to ensure that the dimensionality of the integral result remains consistent.
To model motion blur with 3D-GS, we define the image rendered by 3D-GS at the camera pose at each time $t_i$ as $C_{t_i}(u)$. Therefore, the sharp image $I(u, t_i)$ can be equivalently viewed as the sharp rendering result at different time points, corresponding to $C_{t_i}(u)$. Based on this, we can model the blurry image as the time average of multiple sharp images:
\begin{equation}
B(u) \approx \frac{1}{n} \sum_{i=1}^{n} C_{t_i}(u)
\end{equation}
Were $\{ t_i \}_{i=1}^n$ are the time points uniformly sampled within the exposure interval, corresponding to the camera poses $\{ P(t_i) \}_{i=1}^n$, which are used to dynamically render the sharp image frames $C_{t_i}(u)$. This averaging form serves as a numerical approximation of the temporal integral, allowing us to model motion blur in the real camera imaging process through multi-time image composition within the 3D-GS framework.

Thus, in 3D-GS, $C(u)$ can be viewed as the time-varying sharp image $I(u,t)$, which reflects the scene's observation at a certain moment under different camera poses. The blurry image $B(u)$ is then formed by averaging these sharp images over time, enabling differentiable modeling from Gaussian parameters to blurry images.

\subsection{Event-based Motion Modeling}
An Event Camera is a visual sensor that captures brightness changes with high temporal resolution (typically on the microsecond scale) in an asynchronous manner. Unlike traditional frame cameras, an event camera does not record complete images, but instead triggers events at each pixel when the logarithmic brightness change exceeds a set threshold $\Theta$. 
\begin{equation}
p\Theta = \log I(u, t) - \log I(u, t_{\text{prev}}), \quad \text{where} \quad \Theta > 0
\end{equation}
where \( I(u, t) \) represents the brightness of the pixel at position \( u = (x, y) \) at time \( t \), and \( p \in \{+1, -1\} \) is the polarity, indicating whether the brightness has increased or decreased. Thus, each event can be represented as a quadruple $e = (x, y, t, p)$.

This single event records the pixel change at a certain moment. However, in practical applications, we are more concerned with the flow or temporal sequence of events. An event stream consists of multiple events ordered by time, describing the trend of intensity changes at the pixel location over a period of time. Each event is associated with the previous event through its timestamp \( t \) and pixel position \( (x, y) \).
Given a pixel location $u$ and a time window $[t_{\text{pre}}, t_{\text{cur}}]$, the cumulative event flow $E(t_{\text{pre}}, t_{\text{cur}}, u)$ within this window is defined as:
\begin{equation}
E(t_{\text{pre}}, t_{\text{cur}}, u) = \sum_{s \in [t_{\text{pre}}, t_{\text{cur}}]} p(s, u) 
\end{equation}
where $p(s, u)$ represents the polarity of the event triggered at pixel location $u$ at time $s$. 

According to the event generation model, the cumulative polarity of events is linearly related to the logarithmic brightness difference between images:
\begin{equation}
E(t_{\text{pre}}, t_{\text{cur}}, u) = \frac{\log I(t_{\text{cur}}, u) - \log I(t_{\text{pre}}, u)}{\Theta}
\end{equation}
where $I(t_{\text{pre}}, u)$ and $I(t_{\text{cur}}, u)$ represent the brightness of the latent sharp image at pixel $u$ at times $t_{\text{pre}}$ and $t_{\text{cur}}$, respectively. This equation establishes an explicit relationship between the event stream and the image brightness.

To associate event streams with traditional image representations, the EDI model proposes a method that combines the event sequence with a blurred image to estimate the latent sharp image. Specifically, let the blurred image \( B(u) \) be captured during the camera exposure interval \( [t_s, t_e] \), where \( u = (x, y) \) denotes a pixel location. During the exposure, the camera continuously receives event data that records changes in pixel brightness. To describe this process, we introduce a logarithmic brightness function \( L(u, t) \), defined as:
\begin{equation}
L(u, t) = L(u, t_r) + \int_{t_r}^{t} \Delta L(u, s) \, ds
\end{equation}
where \( t_r \) is a reference time, and \( \Delta L(u, s) \) denotes the brightness change per unit time, which can be approximately reconstructed from the event stream. By substituting this brightness change model into the blurred image expression Eq.~\ref{model_blur}, we obtain:
\begin{equation}
B(u) = \frac{1}{\tau} \int_{t_s}^{t_e} \exp(L(u, t)) \, dt, \quad \text{where } \tau = t_e - t_s
\end{equation}

Further substituting the expression for \( L(u, t) \), we derive a nonlinear integral equation with respect to the latent sharp image at the reference time \( L(u, t_r) \). Let \( I(u, t_r) = \exp(L(u, t_r)) \) represent the latent sharp image, then the entire modeling process can be seen as decomposing the blurred image \( B(u) \) using the event stream \( \Delta L \) to recover \( I(u, t_r) \). By optimizing this model, the EDI method can not only restore the sharp image but also recover the pixel-wise brightness variation trajectory during the exposure, i.e., the motion trace captured by the camera.

In this paper, we construct constraints using event information, designing supervised losses aimed at optimizing the reconstruction quality, starting from both the event generation mechanism and the blurry image formation process.

\subsection{Multi-modal Constraints for Motion Deblurring}
\subsubsection{Image Deblurring Constraint}
Based on the previous modeling of motion blur, the blurry image can be approximated as the average result of multiple latent sharp images at different time points. Therefore, based on the motion blur model during the exposure time, we construct the motion blur loss by approximately reconstructing the blurry image:
\begin{equation}
L_{\text{blur}} = (1 - \lambda_{\text{SSIM}}) L_1 + \lambda_{\text{SSIM}} L_{\text{SSIM}}
\end{equation}
where $L_1$ loss measures the pixel-level intensity difference of the blurry image, ensuring that the reconstructed image is consistent with the real image in terms of brightness distribution; $L_{\text{SSIM}}$ loss measures the structural similarity loss of the blurry image, preserving high-frequency information such as edges and textures. $\lambda_{\text{SSIM}}$ is a weighting factor that balances the contributions of the two losses, and in this paper, it is set to 0.2 to balance pixel-level accuracy and structural fidelity.
\begin{equation}
L_{1} = \frac{1}{N}\sum_{k=1}^{N}|B_{gt}^k(u) - \hat{B_k}(u)|
\end{equation}
where $\hat{B_k}(u)$ is the $k$-th blurry image synthesized by 3D-GS rendering, and $B_{gt}^k(u)$ is the real blurry image captured, $N$ is the total number of blurred images.

\subsubsection{Event Temporal Consistency Constraint}
In order to fully leverage the event data's ability to perceive brightness changes at high temporal resolution, this paper designs a supervised loss function based on event information. Considering the rapid changes in the scene during the exposure time, the event loss between images at a single time point often suffers from insufficient information. Following the approach in \cite{rudnev2022eventnerf}, we introduce a multi-time window random sampling mechanism to establish brightness change constraints between images at different time points.

Specifically, the exposure interval $[t_f - \tau/2, t_f + \tau/2]$ is uniformly discretized into $n$ sub-time windows $\{(t_i, t_{i+1}]\}_{i=0}^{n-1}$, and the event polarities within each sub-window are accumulated to obtain the observed event polarity $E(t_s, t_{i+1}, u)$. Simultaneously, the log-brightness values of the images at each time point are rendered by 3D-GS, and the predicted brightness difference is constructed as:
\begin{equation}
\hat{E}(t_s, t_{i+1}, u) = \frac{1}{\Theta} \left( \log I(t_{i+1}, u) - \log I(t_s, u) \right)
\end{equation}
where $t_s$ is the randomly sampled start time for each sub-window, ensuring that multiple latent image states during the exposure period are covered.

Based on this, the event supervision loss function is defined as:
\begin{equation}
L_{ev} = \frac{1}{n-1} \sum_{i=0}^{n-2} \left\| E(t_s, t_{i+1}, u) - \hat{E}(t_s, t_{i+1}, u) \right\|_1, \text{where }t_s \in [t_0, t_i] 
\end{equation}
where $n$ is the latent number of latent sharp images within the exposure time.
The loss utilizes information from the event stream to provide brightness change details at different moments during the exposure time of a blurred image, thereby offering strong support for eliminating motion blur.

\subsubsection{EDI Prior Constraint}
In this paper, we propose an event loss optimization strategy based on the intermediate moment clarity prior. By leveraging the high temporal resolution and high dynamic range characteristics of the event stream, we jointly optimize the multi-window event loss and motion blur loss. Specifically, the EDI prior is used as the clarity prior at the intermediate moment. We design the following EDI prior loss function:
\begin{equation}
L_{\text{EDI}} = (1 - \lambda_{\text{SSIM}}) L_1 + \lambda_{\text{SSIM}} L_{\text{SSIM}}
\end{equation}
where $L_1$ loss measures the pixel-level intensity difference between the de-blurry image and latent sharp image:
\begin{equation}
L_{1} = \frac{1}{N}\sum_{k=1}^{N}|B_{EDI}^k(u) - \hat{B}{_{sharp}^k}(u)|
\end{equation}
where $\hat{B}{_{sharp}^k}(u)$ is the latent sharp image at intermediate moments during the exposure time 
, and $B_{EDI}^k(u)$ is the de-blurry image computed by EDI.

\subsubsection{Final Loss Function}
In this paper, we propose a comprehensive loss function by integrating RGB images and event stream information to model motion blur. The goal is to improve the accuracy and reconstruction quality of image deblurring through multiple constraints. The final comprehensive loss function can be expressed as:
\begin{equation}
L_{\text{total}} = \lambda_\text{blur}L_{\text{blur}} + \lambda_\text{ev}L_{\text{ev}} + \lambda_\text{EDI}L_{\text{EDI}}
\end{equation}
where \( L_{\text{blur}} \) represents the motion blur loss, \( L_{\text{ev}} \) represents the event-based loss, and \( L_{\text{EDI}} \) represents the EDI prior loss.

\section{Experiments}
\subsection{Dataset and Experimental Setup}
\subsubsection{Dataset}
We have opted to utilize the dataset comprising dual modalities of blurred images and event data, as proposed by Qi et al. \cite{Qi_2024}, for our deblurring experiments. This dataset includes both synthetic and real-world data.
For synthetic data, it encompassed five challenging deblurring scenes, with the simulation of motion blur incorporating scenarios such as camera shake and variations in camera motion speed during exposure, all designed to emulate realistic motion blur effects.
For real-world data, it included two scenes: \textit{Bar} and \textit{Classroom}, captured using the DAVIS-346 color event camera~\cite{davis346}. Both scenes were recorded under low-light conditions using a handheld camera. Unlike EBAD-NeRF \cite{Qi_2024}, which evaluates using only four ground-truth images from specific viewpoints, our experiments incorporated all 28 newly synthesized images from novel viewpoints for the \textit{Bar} scene and all 16 ground-truth images for the \textit{Classroom} scene.
Additionally, we used the deblurred images generated by the EDI algorithm, along with novel view synthesized ground truth images, as inputs to COLMAP to re-estimate camera poses. This process successfully provided initial camera poses and sparse SfM point clouds, which were used to initialize the 3D-GS.

\subsubsection{Experimental Setup}
For motion blur modeling, we adopt the same number of discrete latent sharp images $n$ as EBAD-NeRF~\cite{Qi_2024} to ensure consistency and fair comparison. Specifically, we set $n=5$ for synthetic datasets and $n=7$ for real-world sequences.
Regarding the loss function weight coefficients, we set $\lambda_{\text{blur}} = 1$, $\lambda_{\text{EDI}} = 1$, $\lambda_{\text{ev}} = 0.1$, and $L_{\text{SSIM}}$ is set to 0.2.
To ensure sufficient model convergence and stable performance, we train the network for 20,000 iterations.

\begin{figure*}[htbp]
    \centering
    \begin{minipage}[t]{0.216\textwidth}
        \includegraphics[width=\linewidth]{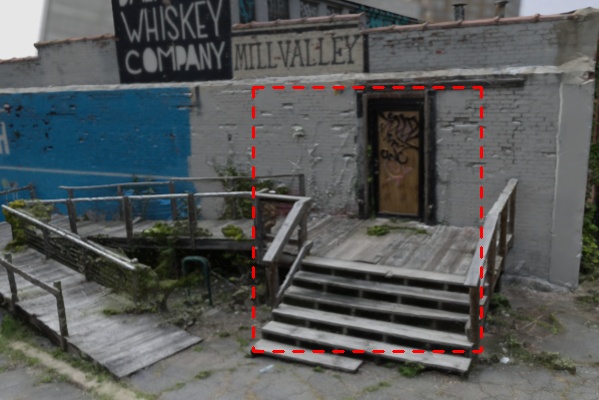}
        \subcaption{Factory}
    \end{minipage}
    \hspace{0.1mm}
    \begin{minipage}[t]{0.125\textwidth}
        \includegraphics[width=\linewidth]{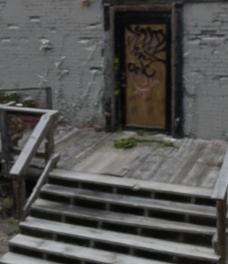}
        \subcaption{GT}
    \end{minipage}
    \begin{minipage}[t]{0.125\textwidth}
        \includegraphics[width=\linewidth]{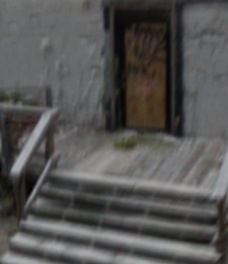}
        \subcaption{$L_{\text{blur}}$}
    \end{minipage}
    \begin{minipage}[t]{0.125\textwidth}
        \includegraphics[width=\linewidth]{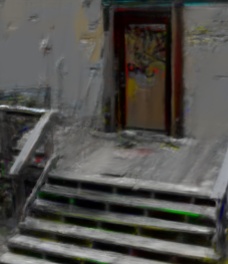}
        \subcaption{$L_{\text{ev}}+L_{\text{EDI}}$}
    \end{minipage}
    \begin{minipage}[t]{0.125\textwidth}
        \includegraphics[width=\linewidth]{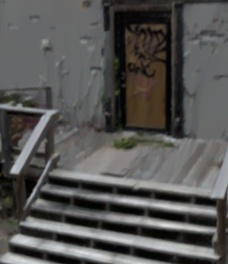}
        \subcaption{$L_{\text{blur}}+L_{\text{ev}}$}
    \end{minipage}
    \begin{minipage}[t]{0.125\textwidth}
        \includegraphics[width=\linewidth]{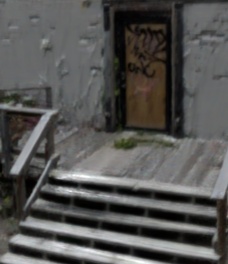}
        \subcaption{$L_{\text{blur}}+L_{\text{EDI}}$}
    \end{minipage}
    \begin{minipage}[t]{0.125\textwidth}
        \includegraphics[width=\linewidth]{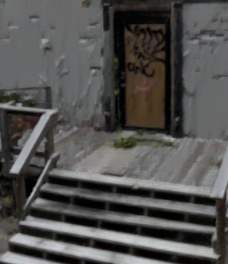}
        \subcaption{$L_{\text{total}}$}
    \end{minipage}
    \caption{Qualitative novel views rendering results of different loss functions on a synthetic dataset.}
    \Description{Qualitative novel views rendering results of different loss functions on synthetic dataset.}
    \label{fig:ablation_syn}
\end{figure*}

\begin{figure*}[htbp]
    \centering
    \begin{minipage}[t]{0.216\textwidth}
        \includegraphics[width=\linewidth]{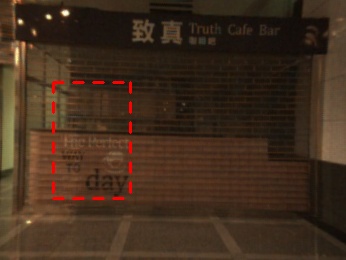}
        \subcaption{Bar}
    \end{minipage}
    \hspace{0.1mm}
    \begin{minipage}[t]{0.125\textwidth}
        \includegraphics[width=\linewidth, height=2.89cm]{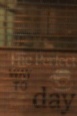}
        \subcaption{GT}
    \end{minipage}
    \begin{minipage}[t]{0.125\textwidth}
        \includegraphics[width=\linewidth, height=2.89cm]{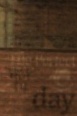}
        \subcaption{$L_{\text{blur}}$}
    \end{minipage}
    \begin{minipage}[t]{0.125\textwidth}
        \includegraphics[width=\linewidth, height=2.89cm]{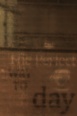}
        \subcaption{$L_{\text{ev}}+L_{\text{EDI}}$}
    \end{minipage}
    \begin{minipage}[t]{0.125\textwidth}
        \includegraphics[width=\linewidth, height=2.89cm]{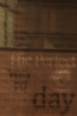}
        \subcaption{$L_{\text{blur}}+L_{\text{ev}}$}
    \end{minipage}
    \begin{minipage}[t]{0.125\textwidth}
        \includegraphics[width=\linewidth, height=2.89cm]{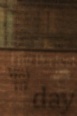}
        \subcaption{$L_{\text{blur}}+L_{\text{EDI}}$}
    \end{minipage}
    \begin{minipage}[t]{0.125\textwidth}
        \includegraphics[width=\linewidth, height=2.89cm]{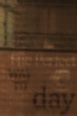}
        \subcaption{$L_{\text{total}}$}
    \end{minipage}
    \caption{Qualitative novel views rendering results of different loss functions on real dataset.}
    \Description{Qualitative novel views rendering results of different loss functions on real dataset.}
    \label{fig:ablation_real}
\end{figure*}

\begin{table*}[htbp]
\renewcommand{\arraystretch}{1.2}
\centering
\setlength{\tabcolsep}{6mm}
\caption{Ablation study of different loss combinations on synthetic and real datasets.}
\label{tab:ablation_combined}
\begin{tabular}{l|ccc|ccc}
\toprule
\multirow{2}{*}{\textbf{Loss Combination}} & \multicolumn{3}{c|}{\textbf{Synthetic Dataset}} & \multicolumn{3}{c}{\textbf{Real Dataset}} \\
& PSNR↑ & SSIM↑ & LPIPS↓ & PSNR↑ & SSIM↑ & LPIPS↓ \\
\midrule
$L_{\text{blur}}$ & 26.89 & 0.803 & 0.258 & 30.24 & 0.858 & 0.148 \\
$L_{\text{blur}} + L_{\text{ev}}$ & 29.19 & 0.874 & 0.159 & 31.08 & 0.902 & 0.162 \\
$L_{\text{blur}} + L_{\text{EDI}}$ & 28.72 & 0.841 & 0.144 & 30.47 & 0.878 & \textbf{0.132} \\
$L_{\text{ev}} + L_{\text{EDI}}$ & 25.00 & 0.814 & 0.231 & 30.04 & 0.897 & 0.176 \\
$L_{\text{total}}$ & \textbf{29.99} & \textbf{0.875} & \textbf{0.119} & \textbf{31.17} & \textbf{0.903} & 0.161 \\
\bottomrule
\end{tabular}
\end{table*}

\subsubsection{Quantitative Evaluation Metrics}
We employ Peak Signal-to-Noise Ratio (PSNR), Structural Similarity Index Measure (SSIM)\cite{image_quality_assessment}, and AlexNet-based Learned Perceptual Patch Similarity (LPIPS) \cite{Zhang_2018_CVPR_LPIPS, alexnet} to evaluate the similarity between rendered images and ground truth images.

\subsection{Ablation Study}
To investigate the collaborative mechanism of the multi-modal loss functions, we conduct a systematic ablation study using a controlled variable approach. Table~\ref{tab:ablation_combined} shows the quantitative impact of different loss combinations on both synthetic and real-world sequences, with results averaged over all evaluated scenes. Figures~\ref{fig:ablation_syn} and~\ref{fig:ablation_real} illustrate the corresponding visual results, clearly showcasing the performance of each method under novel view rendering scenarios.

\subsubsection{Experimental Analysis on Synthetic Dataset}
When only using the motion blur loss $L_{\text{blur}}$, the deblurring performance of the model is relatively poor, with the reconstructed images exhibiting noticeable detail blurring. This is mainly due to the lack of constraints on the intensity changes during exposure. After introducing the event loss $L_{\text{ev}}$, both PSNR and SSIM improve, indicating that the event stream effectively supplements the dynamic information missing from the standard camera and alleviates the loss of high-frequency details. However, the LPIPS remains at 0.159, which is still higher than that of the combination using $L_{\text{EDI}}$, suggesting limited improvement in detail restoration.Replacing $L_{\text{ev}}$ with the event double integral image loss $L_{\text{EDI}}$ reduces the LPIPS to 0.144 and results in sharper edge details. However, the PSNR drops to 28.72, indicating insufficient global luminance consistency.

\subsubsection{Experimental Analysis on Real Dataset}
After combining the three losses, the metrics show the best overall performance. The complementary nature of the three losses is equally effective in real-world scenarios. \(L_{\text{blur}}\) constrains the global motion blur, \(L_{\text{ev}}\) constrains the intensity variation during the exposure time, and \(L_{\text{EDI}}\) enhances edge details. As a result, the fused model achieves the optimal reconstruction quality in real scenes. Although the LPIPS score slightly increases compared to the combination of \(L_{\text{blur}} + L_{\text{EDI}}\), this small fluctuation is acceptable, and the overall performance still outperforms the other combinations.

\begin{figure*}[htbp]
    \centering
    \begin{tabular}{@{}cc@{}}
        \raisebox{-\totalheight}{
            \begin{minipage}[b]{0.28\textwidth}
                \centering
                \includegraphics[width=\linewidth]{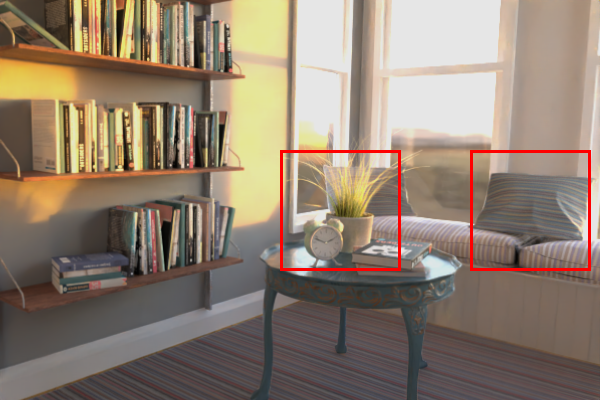}
                \vspace*{-6.5mm}
                \caption*{GT (Synthetic Scene – Cozyroom)}
            \end{minipage}}
        &
        \raisebox{-\totalheight}{
            \begin{minipage}[b]{0.70\textwidth}
                \begin{subfigure}[b]{0.13\textwidth}
                    \includegraphics[width=\linewidth]{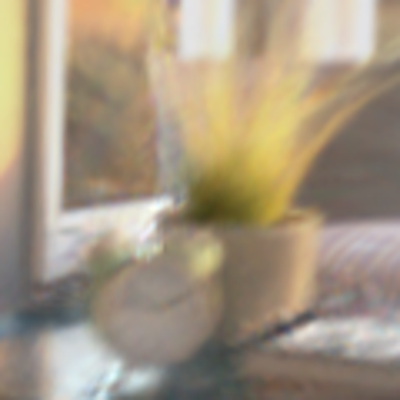}
                \end{subfigure}\hfill
                \begin{subfigure}[b]{0.13\textwidth}
                    \includegraphics[width=\linewidth]{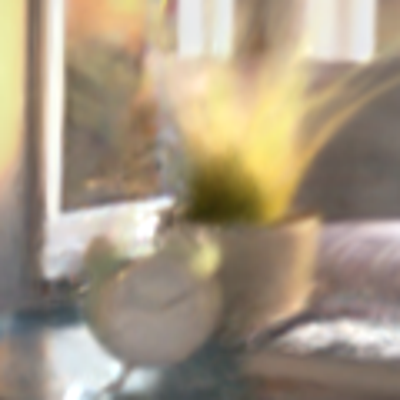}
                \end{subfigure}\hfill
                \begin{subfigure}[b]{0.13\textwidth}
                    \includegraphics[width=\linewidth]{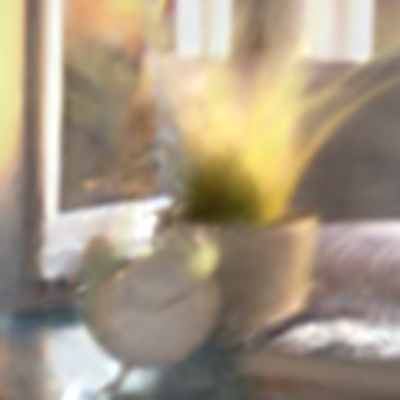}
                \end{subfigure}\hfill
                \begin{subfigure}[b]{0.13\textwidth}
                    \includegraphics[width=\linewidth]{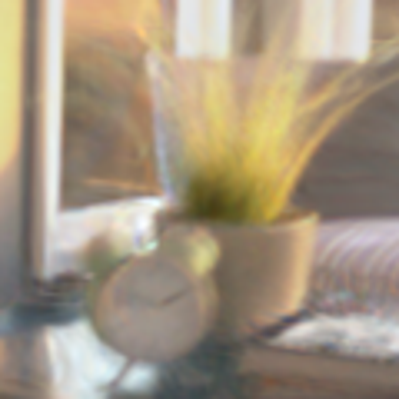}
                \end{subfigure}\hfill
                \begin{subfigure}[b]{0.13\textwidth}
                    \includegraphics[width=\linewidth]{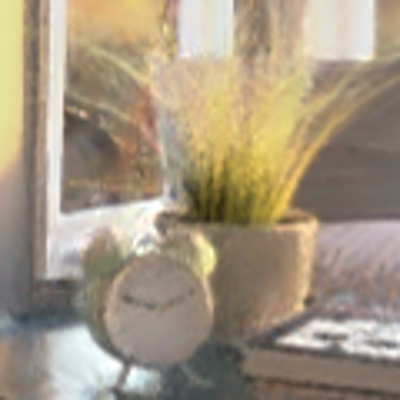}
                \end{subfigure}\hfill
                \begin{subfigure}[b]{0.13\textwidth}
                    \includegraphics[width=\linewidth]{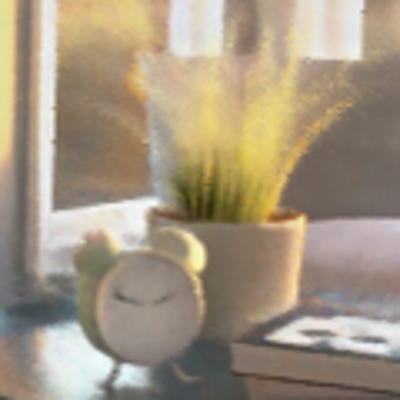}
                \end{subfigure}\hfill
                \begin{subfigure}[b]{0.13\textwidth}
                    \includegraphics[width=\linewidth]{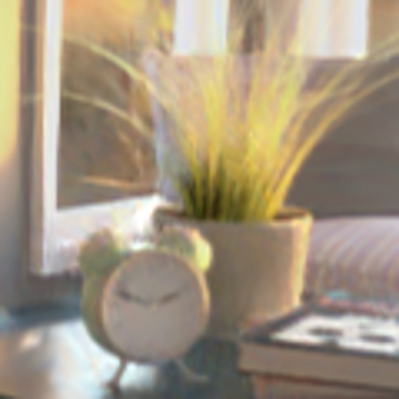}
                \end{subfigure}
                
                \vspace{1mm}
                
                \begin{subfigure}[b]{0.13\textwidth}
                    \includegraphics[width=\linewidth]{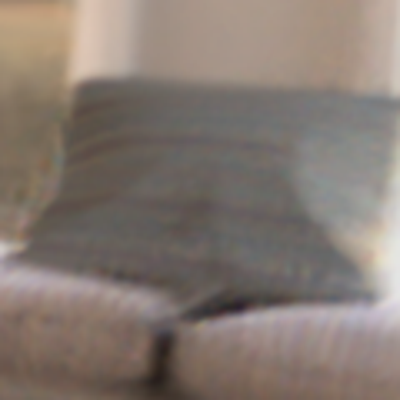}
                    \caption*{3DGS}
                \end{subfigure}\hfill
                \begin{subfigure}[b]{0.13\textwidth}
                    \includegraphics[width=\linewidth]{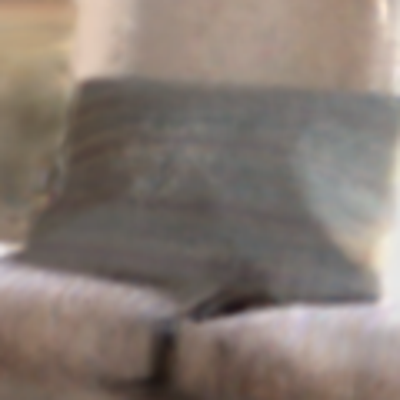}
                    \caption*{3DGS+MPR}
                \end{subfigure}\hfill
                \begin{subfigure}[b]{0.13\textwidth}
                    \includegraphics[width=\linewidth]{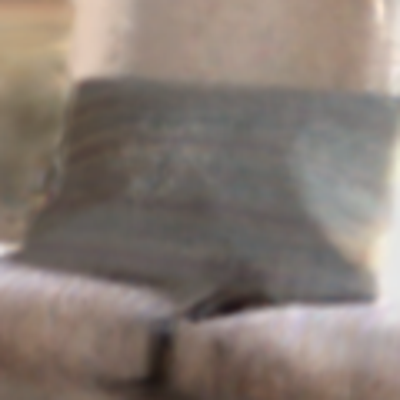}
                    \caption*{BAD-GS}
                \end{subfigure}\hfill
                \begin{subfigure}[b]{0.13\textwidth}
                    \includegraphics[width=\linewidth]{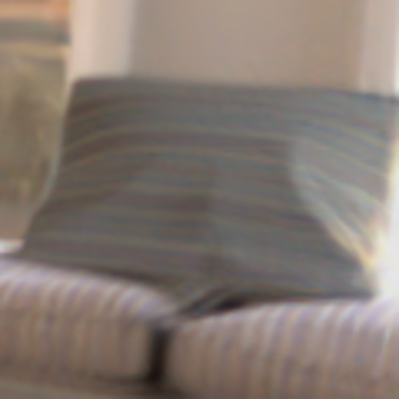}
                    \caption*{BAD-GS*}
                \end{subfigure}\hfill
                \begin{subfigure}[b]{0.13\textwidth}
                    \includegraphics[width=\linewidth]{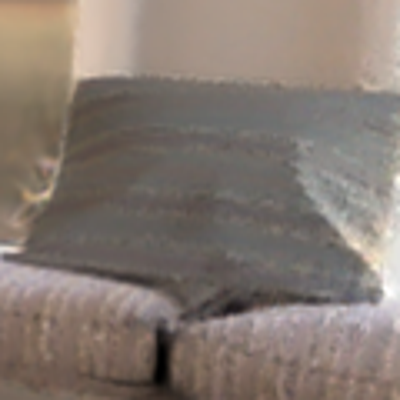}
                    \caption*{3DGS+EDI}
                \end{subfigure}\hfill
                \begin{subfigure}[b]{0.13\textwidth}
                    \includegraphics[width=\linewidth]{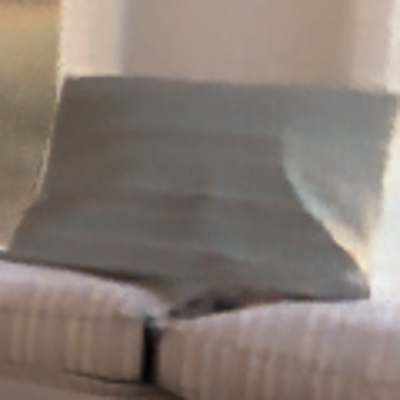}
                    \caption*{EBAD-NeRF}
                \end{subfigure}\hfill
                \begin{subfigure}[b]{0.13\textwidth}
                    \includegraphics[width=\linewidth]{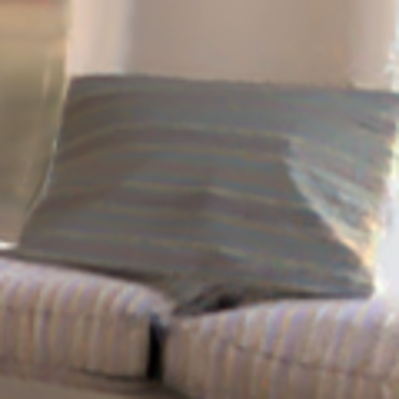}
                    \caption*{Ours}
                \end{subfigure}
            \end{minipage}}
        
        \\[1mm] 
        
        \raisebox{-\totalheight}{
            \begin{minipage}[b]{0.28\textwidth}
                \centering
                \includegraphics[width=\linewidth]{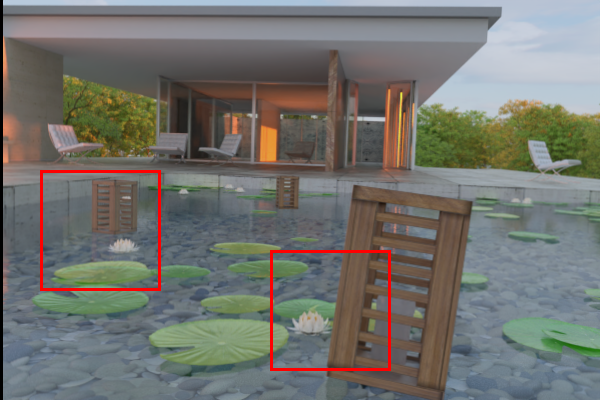}
                \vspace*{-6.5mm}
                \caption*{GT (Synthetic Scene – Pool)} 
            \end{minipage}}
        &
        \raisebox{-\totalheight}{
            \begin{minipage}[b]{0.70\textwidth}
                \begin{subfigure}[b]{0.13\textwidth}
                    \includegraphics[width=\linewidth]{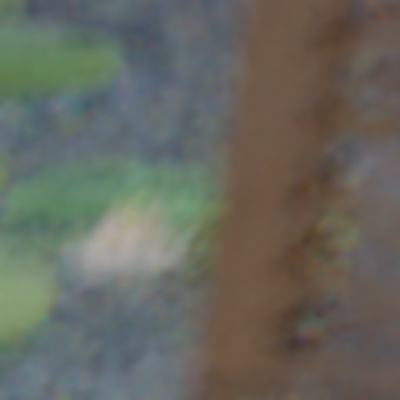}
                \end{subfigure}\hfill
                \begin{subfigure}[b]{0.13\textwidth}
                    \includegraphics[width=\linewidth]{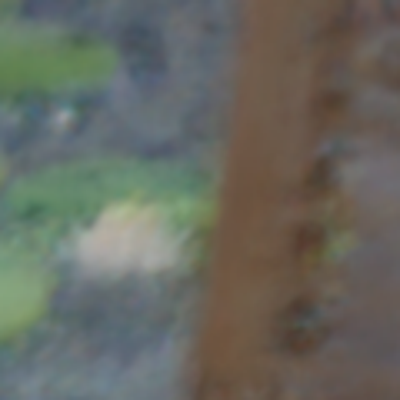}
                \end{subfigure}\hfill
                \begin{subfigure}[b]{0.13\textwidth}
                    \includegraphics[width=\linewidth]{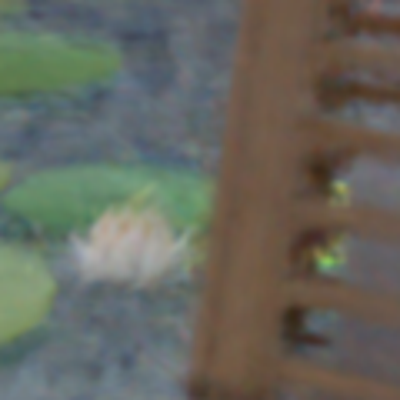}
                \end{subfigure}\hfill
                \begin{subfigure}[b]{0.13\textwidth}
                    \includegraphics[width=\linewidth]{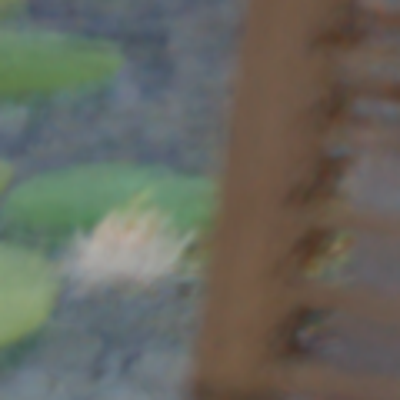}
                \end{subfigure}\hfill
                \begin{subfigure}[b]{0.13\textwidth}
                    \includegraphics[width=\linewidth]{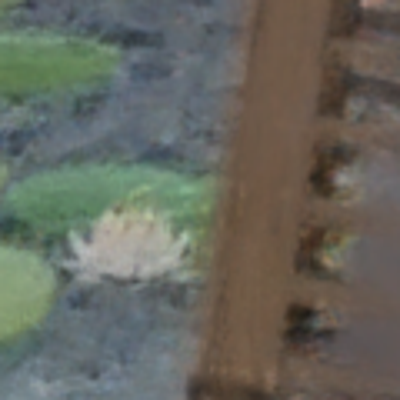}
                \end{subfigure}\hfill
                \begin{subfigure}[b]{0.13\textwidth}
                    \includegraphics[width=\linewidth]{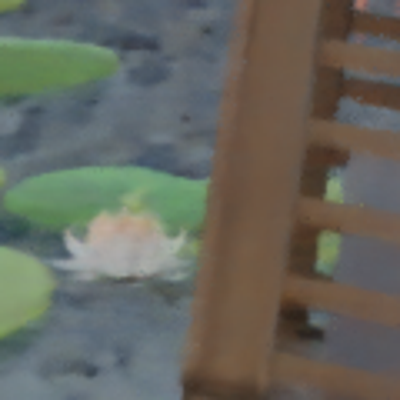}
                \end{subfigure}\hfill
                \begin{subfigure}[b]{0.13\textwidth}
                    \includegraphics[width=\linewidth]{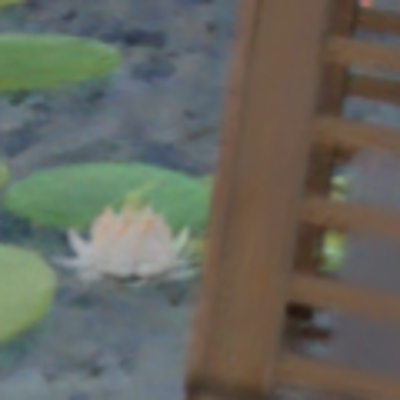}
                \end{subfigure}
                
                \vspace{1mm}
                
                \begin{subfigure}[b]{0.13\textwidth}
                    \includegraphics[width=\linewidth]{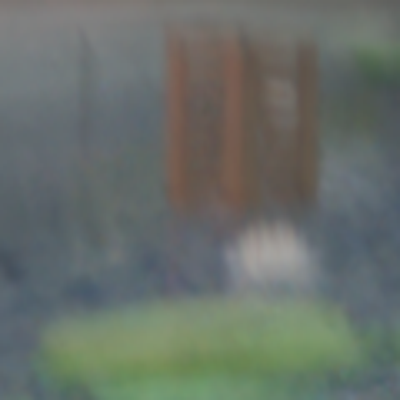}
                    \caption*{3D-GS}
                \end{subfigure}\hfill
                \begin{subfigure}[b]{0.13\textwidth}
                    \includegraphics[width=\linewidth]{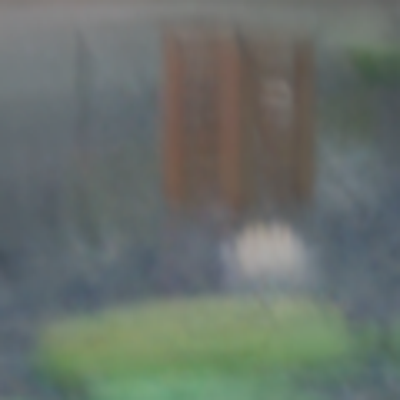}
                    \caption*{3DGS+MPR}
                \end{subfigure}\hfill
                \begin{subfigure}[b]{0.13\textwidth}
                    \includegraphics[width=\linewidth]{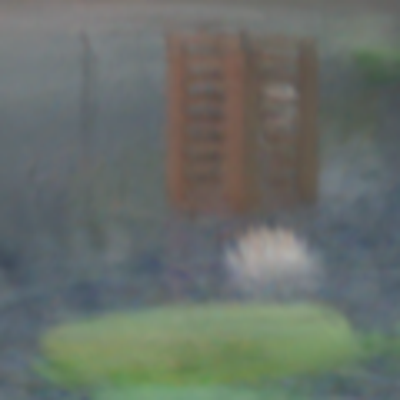}
                    \caption*{BAD-GS}
                \end{subfigure}\hfill
                \begin{subfigure}[b]{0.13\textwidth}
                    \includegraphics[width=\linewidth]{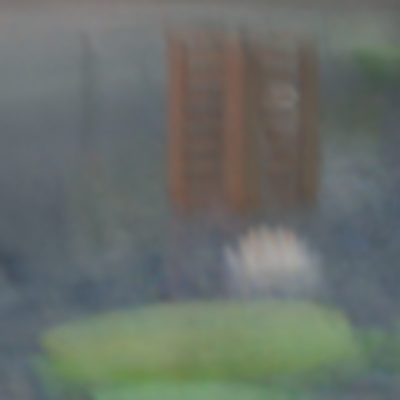}
                    \caption*{BAD-GS*}
                \end{subfigure}\hfill
                \begin{subfigure}[b]{0.13\textwidth}
                    \includegraphics[width=\linewidth]{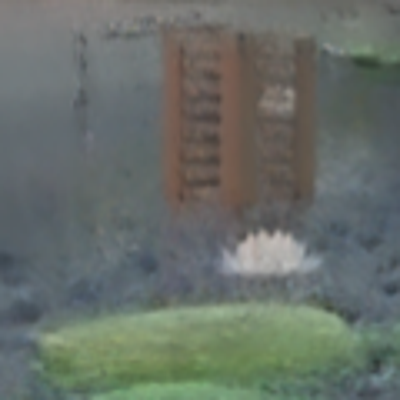}
                    \caption*{3DGS+EDI}
                \end{subfigure}\hfill
                \begin{subfigure}[b]{0.13\textwidth}
                    \includegraphics[width=\linewidth]{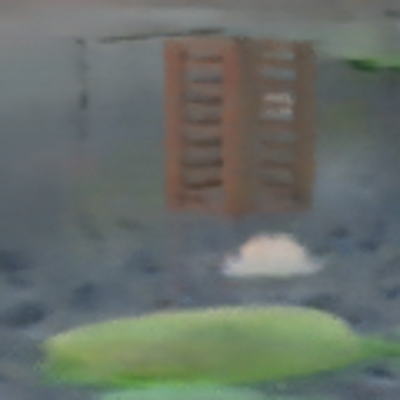}
                    \caption*{EBAD-NeRF}
                \end{subfigure}\hfill
                \begin{subfigure}[b]{0.13\textwidth}
                    \includegraphics[width=\linewidth]{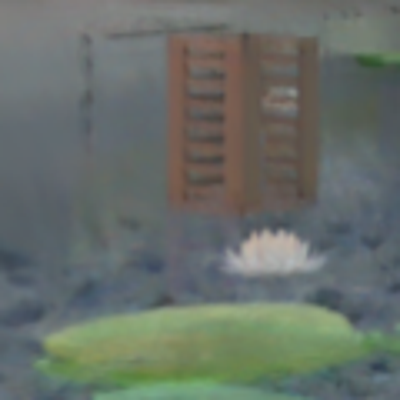}
                    \caption*{Ours}
                \end{subfigure}
            \end{minipage}}
        \\[1mm]
        
        \raisebox{-\totalheight}{
            \begin{minipage}[b]{0.28\textwidth}
                \centering
                \includegraphics[width=\linewidth]{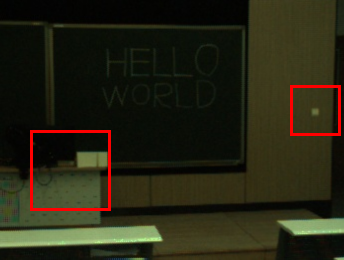}
                \vspace*{-6.5mm}
                \caption*{GT (Real Scene – Class)}
            \end{minipage}}
        &
        \raisebox{-\totalheight}{
            \begin{minipage}[b]{0.70\textwidth}
                \begin{subfigure}[b]{0.13\textwidth}
                    \includegraphics[width=\linewidth]{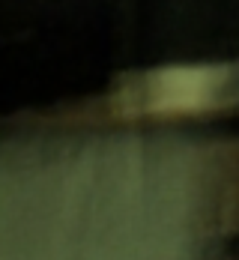}
                \end{subfigure}\hfill
                \begin{subfigure}[b]{0.13\textwidth}
                    \includegraphics[width=\linewidth]{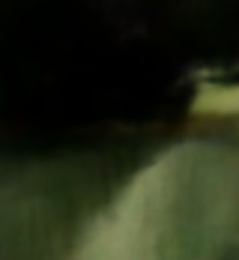}
                \end{subfigure}\hfill
                \begin{subfigure}[b]{0.13\textwidth}
                    \includegraphics[width=\linewidth]{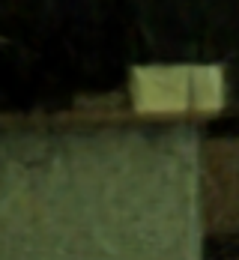}
                \end{subfigure}\hfill
                \begin{subfigure}[b]{0.13\textwidth}
                    \includegraphics[width=\linewidth]{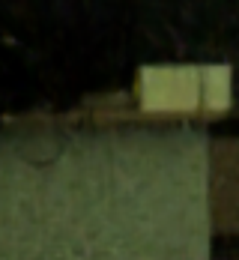}
                \end{subfigure}\hfill
                \begin{subfigure}[b]{0.13\textwidth}
                    \includegraphics[width=\linewidth]{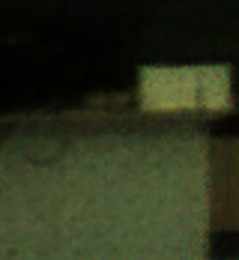}
                \end{subfigure}\hfill
                \begin{subfigure}[b]{0.13\textwidth}
                    \includegraphics[width=\linewidth]{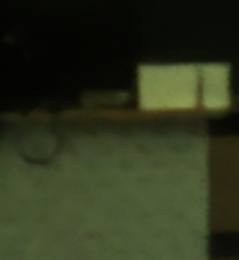}
                \end{subfigure}\hfill
                \begin{subfigure}[b]{0.13\textwidth}
                    \includegraphics[width=\linewidth]{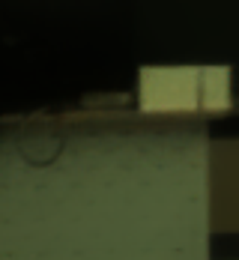}
                \end{subfigure}
                
                \vspace{1mm}
                
                \begin{subfigure}[b]{0.13\textwidth}
                    \includegraphics[width=\linewidth]{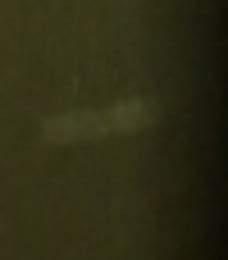}
                    \caption*{3D-GS}
                \end{subfigure}\hfill
                \begin{subfigure}[b]{0.13\textwidth}
                    \includegraphics[width=\linewidth]{fig/3dgs_mpr_c_ROI_0.png}
                    \caption*{3DGS+MPR}
                \end{subfigure}\hfill
                \begin{subfigure}[b]{0.13\textwidth}
                    \includegraphics[width=\linewidth]{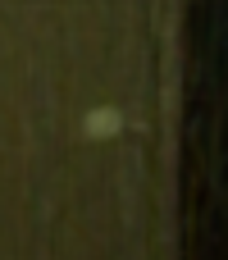}
                    \caption*{BAD-GS}
                \end{subfigure}\hfill
                \begin{subfigure}[b]{0.13\textwidth}
                    \includegraphics[width=\linewidth]{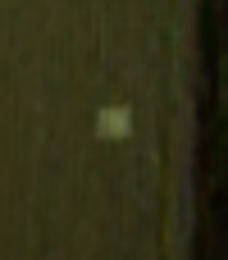}
                    \caption*{BAD-GS*}
                \end{subfigure}\hfill
                \begin{subfigure}[b]{0.13\textwidth}
                    \includegraphics[width=\linewidth]{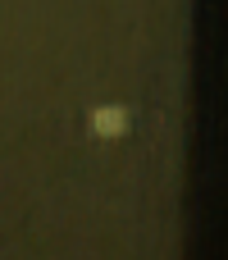}
                    \caption*{3DGS+EDI}
                \end{subfigure}\hfill
                \begin{subfigure}[b]{0.13\textwidth}
                    \includegraphics[width=\linewidth]{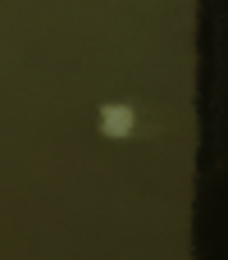}
                    \caption*{EBAD-NeRF}
                \end{subfigure}\hfill
                \begin{subfigure}[b]{0.13\textwidth}
                    \includegraphics[width=\linewidth]{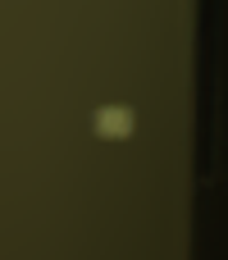}
                    \caption*{Ours}
                \end{subfigure}
            \end{minipage}}
    \end{tabular}
    \caption{Qualitative novel view synthesis results of different methods.}
    \Description{Qualitative novel view synthesis results of different methods on synthetic dataset.}
    \label{fig:dual_gt_comparison}
\end{figure*}

\begin{table*}
\centering
\renewcommand{\arraystretch}{1.1}
\setlength{\tabcolsep}{8pt}
\caption{Comparison of Different Methods on Synthetic Dataset}
\label{tab:method_comparsion_syn}
\begin{tabular}{@{}llccccccc@{}}
\toprule
\textbf{Scene} & \textbf{Metric} & 3D-GS & 3D-GS+MPR & BAD-GS & BAD-GS* & 3D-GS+EDI & EBAD-NeRF & \textbf{Our Method} \\
\midrule
Cozyroom & PSNR↑ & 27.84 & 26.61 & {31.12} & 29.53 & 30.19 & 30.60 & \textbf{32.68} \\
         & SSIM↑ & 0.855 & 0.840 & {0.930} & 0.905 & 0.898 & \textbf{0.948} & 0.935 \\
         & LPIPS↓ & 0.222 & 0.239 & {0.084} & 0.130 & 0.104 & 0.114 & \textbf{0.067} \\
\midrule
Factory  & PSNR↑ & 20.30 & 21.38 & 21.23 & 22.33 & 25.46 & {28.25} & \textbf{28.79} \\
         & SSIM↑ & 0.539 & 0.590 & 0.628 & 0.680 & 0.788 & \textbf{0.915} & 0.865 \\
         & LPIPS↓ & 0.486 & 0.434 & 0.342 & 0.339 & 0.178 & {0.158} & \textbf{0.129} \\
\midrule
Pool     & PSNR↑ & 27.75 & 27.65 & 29.19 & 29.15 & 30.15 & {31.62} & \textbf{32.92} \\
         & SSIM↑ & 0.733 & 0.742 & 0.787 & 0.797 & 0.807 & \textbf{0.924} & 0.881 \\
         & LPIPS↓ & 0.363 & 0.350 & 0.228 & 0.257 & 0.175 & {0.156} & \textbf{0.106} \\
\midrule
Tanabata & PSNR↑ & 19.46 & 19.57 & 20.49 & 20.76 & 24.09 & {25.45} & \textbf{26.93} \\
         & SSIM↑ & 0.570 & 0.592 & 0.669 & 0.670 & 0.670 & \textbf{0.890} & 0.838 \\
         & LPIPS↓ & 0.489 & 0.435 & 0.351 & 0.382 & 0.240 & {0.206} & \textbf{0.151} \\
\midrule
Wine     & PSNR↑ & 20.32 & 20.35 & 23.36 & 21.33 & 25.35 & {26.72} & \textbf{28.67} \\
         & SSIM↑ & 0.564 & 0.564 & 0.732 & 0.650 & 0.755 & \textbf{0.889} & 0.852 \\
         & LPIPS↓ & 0.496 & 0.479 & 0.290 & 0.363 & 0.246 & {0.196} & \textbf{0.140} \\
\midrule
Average  & PSNR↑ & 23.13 & 23.11 & 25.08 & 24.62 & 27.05 & {28.53} & \textbf{30.00} \\
         & SSIM↑ & 0.652 & 0.665 & 0.749 & 0.740 & 0.784 & \textbf{0.913} & 0.874 \\
         & LPIPS↓ & 0.411 & 0.387 & 0.259 & 0.294 & 0.189 & {0.166} & \textbf{0.119} \\
\bottomrule
\end{tabular}
\end{table*}

\begin{table*}
\centering
\setlength{\tabcolsep}{8pt}
\caption{Comparison of Different Methods on Real Dataset}
\label{tab:method_comparison_real}
\begin{tabular}{@{}llccccccc@{}}
\toprule
\textbf{Scene} & \textbf{Metric} & 3D-GS & 3D-GS+MPR & BAD-GS & BAD-GS* & 3D-GS+EDI & EBAD-NeRF & \textbf{Our Method} \\
\midrule
Bar     & PSNR↑ & 25.07 & 24.23 & 28.08 & 28.03 & 28.16 & 28.03 & \textbf{29.87} \\
        & SSIM↑ & 0.744 & 0.731 & 0.781 & 0.780 & 0.774 & 0.807 & \textbf{0.880} \\
        & LPIPS↓ & 0.442 & 0.475 & 0.226 & 0.245 & 0.259 & 0.209 & \textbf{0.221} \\
\midrule
Classroom & PSNR↑ & 23.37 & 23.07 & 30.64 & 31.51 & 27.86 & 30.72 & \textbf{32.47} \\
          & SSIM↑ & 0.751 & 0.751 & 0.857 & 0.879 & 0.852 & 0.908 & \textbf{0.926} \\
          & LPIPS↓ & 0.359 & 0.344 & 0.137 & 0.110 & 0.148 & 0.121 & \textbf{0.102} \\
\midrule
Average  & PSNR↑ & 24.22 & 23.65 & 29.36 & 29.77 & 28.01 & 29.37 & \textbf{31.17} \\
         & SSIM↑ & 0.748 & 0.741 & 0.819 & 0.830 & 0.813 & 0.857 & \textbf{0.903} \\
         & LPIPS↓ & 0.400 & 0.409 & 0.182 & 0.178 & 0.203 & 0.165 & \textbf{0.161} \\
\bottomrule
\end{tabular}
\end{table*}

\subsection{Comparison with Existing Methods}
\subsubsection{Experimental Comparison on Synthetic Dataset}
This experiment comprehensively compares the quantitative results of different methods on the new viewpoint synthesis task for synthetic sequences (as shown in Table \ref{tab:method_comparsion_syn}), and also provides the results of new viewpoint rendering (Figure \ref{fig:dual_gt_comparison}) as qualitative evaluation references.
First, the 3D-GS method directly utilizes blurred images for reconstruction, which results in noticeable blurring in the new viewpoint rendering results. Quantitative results show that this method performs relatively poorly in terms of image quality, structural similarity, and perceptual quality, requiring improvement. Secondly, the 3D-GS+MPRNet method achieves some improvements by replacing the blurred images with those deblurred by the MPRNet\cite{Zamir2021MPRNet} module. The results show that SSIM increases to 0.665, indicating that the structural similarity of the image has improved. However, PSNR and LPIPS slightly decrease, suggesting that MPRNet still has limited effectiveness in handling severe motion blur scenarios. In contrast, BAD-GS\cite{ye2024gsplatopensourcelibrarygaussian} significantly improves the indicators and shows strong performance in motion blur modeling. Furthermore, the BAD-GS cubic method, which uses bilinear B-spline interpolation for trajectory recovery on top of BAD-GS, results in worse performance. In our figures and tables, we denote this variant as BAD-GS* for clarity. Nevertheless, its performance is still better than 3D-GS and 3D-GS+MPRNet, indicating that bilinear B-spline interpolation has a limited impact on the synthetic dataset and does not significantly degrade performance. The 3D-GS+EDI method, using the event-driven deblurring algorithm EDI, significantly improves image quality, demonstrating the effectiveness of event information. The EBAD-NeRF method, by combining motion blur modeling with event flow supervision of light intensity changes during exposure and optimizing the camera pose over the exposure time in a neural radiance field, shows near-optimal performance. However, the rendering speed of EBAD-NeRF is limited by the NeRF query MLP, preventing real-time rendering.
Finally, our proposed method performs the best among all the tested methods. Although the SSIM of 0.871 is slightly lower than EBAD-NeRF, it performs outstandingly in PSNR and LPIPS, the two key indicators. More importantly, our method supports real-time rendering, which gives it a significant advantage in practical applications.

\subsubsection{Experimental Comparison on Real Dataset}
In this experiment, we conducted a comparative analysis of different methods for the new view synthesis task on real sequences, and provided quantitative evaluation results (as shown in Table \ref{tab:method_comparison_real}) and qualitative evaluation results (as shown in Figure \ref{fig:dual_gt_comparison}).
Firstly, the 3D-GS method performs relatively poorly on real sequences due to the blur present in the input images. Quantitative results show that the image quality, structural similarity, and perceptual quality in real scenes are at moderate levels, with considerable room for improvement. Secondly, the 3D-GS+MPRNet method shows a decline in quantitative results. This may be because the MPRNet module lacks generalization ability in handling the complex blur situations in real scenes, leading to no significant improvement in overall performance. In contrast, the BAD-GS method shows strong performance on real sequences, indicating that the method can effectively model the blurry physical processes in real scenes, thus achieving significant improvements in image quality, structural similarity, and perceptual quality. Furthermore, the BAD-GS* method further enhances performance. PSNR reaches 29.77, SSIM increases to 0.830, and LPIPS decreases to 0.178, indicating that the bilinear B-spline interpolation method can effectively optimize trajectory recovery in real sequences, further improving overall performance. The 3D-GS+EDI method achieves a good performance on real sequences by using the EDI deblurring algorithm. Although the performance is slightly lower than BAD-GS and BAD-GS*, it still outperforms 3D-GS and 3D-GS+MPRNet. The results indicate that event-driven deblurring algorithms have good deblurring effects in real scenes. However, because this method does not simultaneously optimize camera poses and scene representation, its performance still has some limitations. The EBAD-NeRF method also demonstrates strong performance on real datasets. While it excels in terms of image quality and structural similarity, its performance on the LPIPS metric still leaves room for improvement. The discrepancy in the metrics reported for the EBAD-NeRF method compared to the original paper arises from the use of real datasets with more viewpoints in our experiments. Quantitative results show that our proposed method performs the best on real sequences, effectively improving image reconstruction in real scenes.

\section{Conclusion}
In summary, the proposed \textbf{EBAD-Gaussian} method fully leverages the complementary advantages of event streams and images, effectively addressing the reconstruction quality limitations of 3D-GS when dealing with severe motion blur. Based on a physically grounded motion blur model, we introduce three types of key supervisory signals: motion blur constraints to ensure baseline reconstruction quality, event stream supervision to recover high-frequency details, and the EDI prior to enhance physical consistency. On this basis, our framework jointly optimizes the Gaussian parameters and the camera poses during exposure, thereby achieving high-fidelity 3D scene reconstruction. Experimental results on both synthetic and real datasets validate the effectiveness of EBAD-Gaussian in camera pose estimation and scene recovery.

\bibliographystyle{unsrt} %
\bibliography{ref}
\end{document}